\documentclass[letterpaper]{article} 
\usepackage{aaai2026}  
\usepackage{times}  
\usepackage{helvet}  
\usepackage{courier}  
\usepackage[hyphens]{url}  
\usepackage{graphicx} 
\usepackage{natbib}  
\usepackage{caption} 
\usepackage{algorithm}
\usepackage{algpseudocode}
\usepackage{amsmath}   
\usepackage{amssymb}   
\usepackage{multirow}
\usepackage[table,xcdraw]{xcolor}
\usepackage{graphicx}     
\usepackage{caption}

\usepackage{newfloat}
\usepackage{listings}
\DeclareCaptionStyle{ruled}{labelfont=normalfont,labelsep=colon,strut=off} 
\floatstyle{ruled}
\newfloat{listing}{tb}{lst}{}
\floatname{listing}{Listing}
\title{FantasyStyle: Controllable Stylized Distillation for 3D Gaussian Splatting}
\author{
    Yitong Yang\textsuperscript{\rm 1},
    Yinglin Wang\textsuperscript{\rm 1}\thanks{Corresponding author.},
    Changshuo Wang\textsuperscript{\rm 3},
    Huajie Wang\textsuperscript{\rm 4, 5},
    Shuting He\textsuperscript{\rm 1, 2}\footnotemark[1]
}
\affiliations{
    \textsuperscript{\rm 1}School of Computing and Artificial Intelligence, Shanghai University of Finance and Economics, Shanghai, China

   \textsuperscript{\rm 2}MoE Key Laboratory of Interdisciplinary Research of Computation and Economics, Shanghai, China\\
    \textsuperscript{\rm 3}Department of Computer Science University College London, London, United Kingdom\\
    \textsuperscript{\rm 4}Shandong University of Finance and Economics, Shandong, China\\
    \textsuperscript{\rm 5}Jinan Yunwei Software Technology Co., Ltd, Shandong, China\\
    yangyitong@stu.sufe.edu.cn, wang.yinglin@shufe.edu.cn, wangchangshuo1@gmail.com, wanghuajie@sdufe.edu.cn, shuting.he@sufe.edu.cn
}

\usepackage{bibentry}

\begin{document}

\maketitle

\begin{abstract}
The success of 3DGS in generative and editing applications has sparked growing interest in 3DGS-based style transfer. However, current methods still face two major challenges: (1) multi-view inconsistency often leads to style conflicts, resulting in appearance smoothing and distortion; and (2) heavy reliance on VGG features, which struggle to disentangle style and content from style images, often causing content leakage and excessive stylization. To tackle these issues, we introduce \textbf{FantasyStyle}, a 3DGS-based style transfer framework, and the first to rely entirely on diffusion model distillation. It comprises two key components: (1) \textbf{Multi-View Frequency Consistency}. We enhance cross-view consistency by applying a 3D filter to multi-view noisy latent, selectively reducing low-frequency components to mitigate stylized prior conflicts. (2) \textbf{Controllable Stylized Distillation}. To suppress content leakage from style images, we introduce negative guidance to exclude undesired content. In addition, we identify the limitations of Score Distillation Sampling and Delta Denoising Score in 3D style transfer and remove the reconstruction term accordingly. Building on these insights, we propose a controllable stylized distillation that leverages negative guidance to more effectively optimize the 3D Gaussians. Extensive experiments demonstrate that our method consistently outperforms state-of-the-art approaches, achieving higher stylization quality and visual realism across various scenes and styles. The code is available at \url{https://github.com/yangyt46/FantasyStyle}.
\end{abstract}    
\section{Introduction}
\label{sec:intro}
With the widespread adoption of 3D content~\cite{he2025survey,ding2025mevis,wang2025point, wang2025taylor} in virtual reality (VR), augmented reality (AR), and related fields, user expectations for visual quality have shifted from traditional photorealism toward more artistic and personalized styles. In this context, style transfer has emerged as a key technique for altering the visual appearance of 3D content, attracting increasing attention from the research community. Previous works~\cite{zhang2022arf,jung2024geometry,zhang2024coarf,zhang2023ref} have explored 3D style transfer based on Neural Radiance Fields (NeRF)~\cite{mildenhall2021nerf}, typically by introducing VGG-based stylization losses to ensure both style consistency and structural coherence in rendered views. However, NeRF suffers from slow rendering and long training times, limiting its practical applications. In contrast, 3D Gaussian Splatting (3DGS)~\cite{kerbl20233d}, as an emerging 3D representation, has become a research focus for 3D style transfer due to its faster rendering and superior visual quality.

\begin{figure}[!t]
    \centering
    \includegraphics[width=\linewidth]{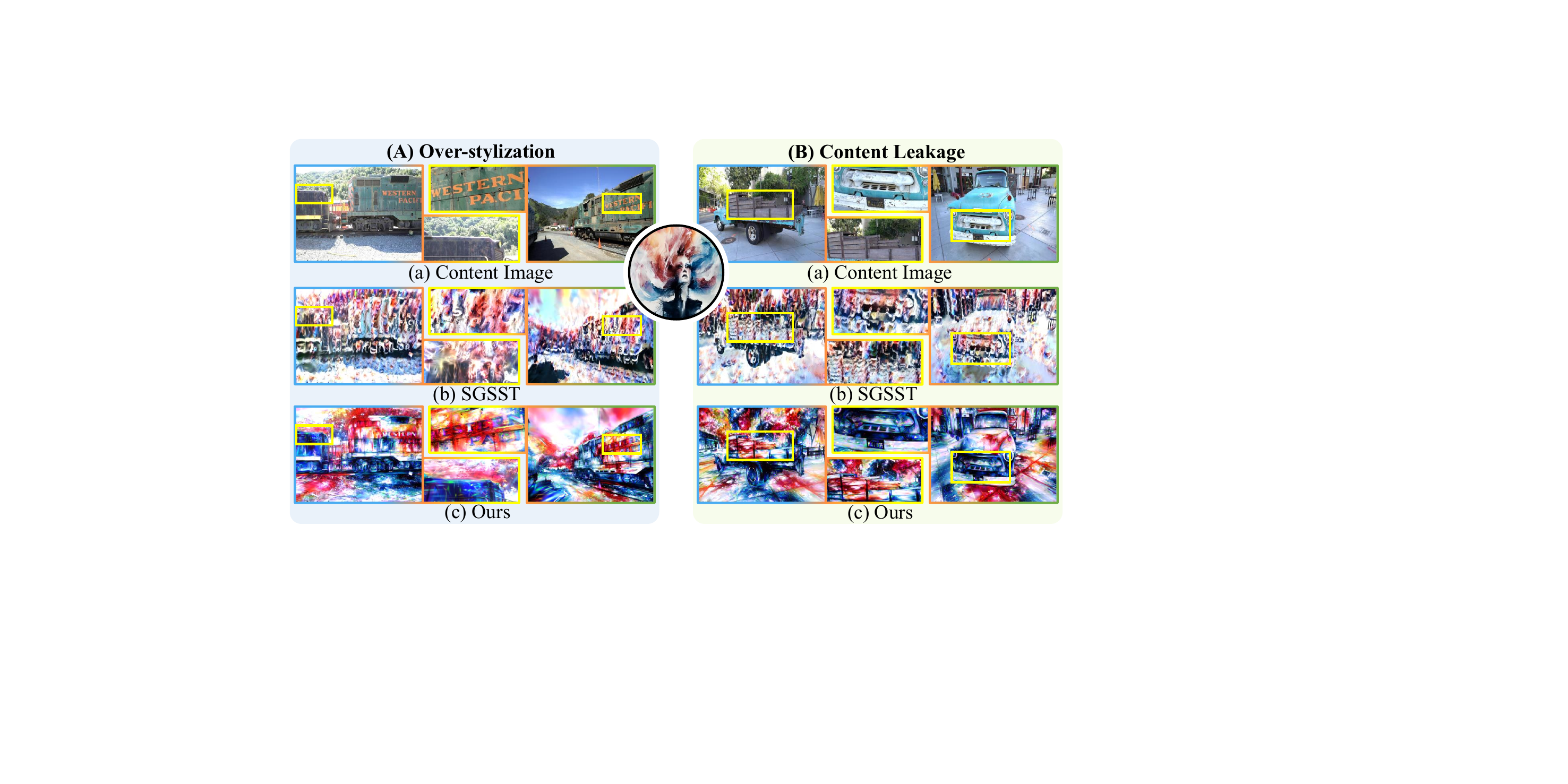}
    \caption{Limitations of the previous works. VGG-based methods often cause over-stylization and content leakage from the style image. In contrast, our approach ensures faithful style transfer and content preservation.}
    \label{fig:teaser}
    \vspace{-10pt}
\end{figure}

Most recent studies on style transfer for 3DGS are based on VGG features, drawing inspiration from the early work on feature statistic matching~\cite{gatys2016image}. These methods~\cite{zhang2024stylizedgs,saroha2024gaussian,kovacs2024g} typically extract VGG features and compute Gram matrices to achieve texture style transfer. Meanwhile, some studies~\cite{zuo2024towards,ArtNVG} attempt to leverage diffusion models to generate stylized 2D images, which are then further optimized back into the 3D scene to enable style transfer. Although these methods have shown promising results, they still face two critical challenges. (1) \textbf{Multi-view inconsistency.} Using 2D diffusion priors to guide 3D stylization often leads to inconsistent appearances across views, especially between adjacent ones. These inconsistencies can lead to style conflicts, hinder optimization, and distort geometry and appearance, as shown in Fig.~\ref{fig:multi-view} (b). (2) \textbf{Content leakage from style images and over-stylization.} As shown in Fig.~\ref{fig:teaser} (A) (B), VGG-based methods struggle to separate content and style, often causing unintended content transfer from the style image. Their emphasis on low-level textures often results in over-stylization, which obscures structural details.

To address the aforementioned challenges, we propose \textbf{FantasyStyle}, a 3DGS-based style transfer framework. Drawing inspiration from FreeU~\cite{si2024freeu} and FreeInit~\cite{wu2024freeinit}, which emphasize the critical role of frequency components in image and video generation, we perform a frequency-domain analysis of 2D multi-view stylized priors. We observe that low-frequency components primarily reflect view-dependent local details and tend to exhibit poor consistency across different viewpoints. In contrast, high-frequency components more stably capture texture features and demonstrate better consistency across views. Therefore, appropriately attenuating the low-frequency components can help alleviate inter-view discrepancies and improve overall multi-view consistency. Based on this observation, we introduce \textbf{Multi-View Frequency Consistency (MVFC)}, which applies a 3D frequency-domain filter to the multi-view latent after DDIM noise injection. MVFC preserves all high-frequency components and selectively suppresses low-frequency components, effectively reducing stylized prior conflicts and improving both consistency and stylization quality.

In parallel, to eliminate potential content leakage from the style image, we incorporate negative guidance during the denoising phase, ensuring that the derived 2D stylized prior remains free of content information. While methods~\cite{chung2023luciddreamer} such as Score Distillation Sampling (SDS) and Delta Denoising Score (DDS) have been employed for 3D scene optimization, they are not directly applicable in our context, as they tend to produce overly smooth images that lose important brushstroke details present in style images. To overcome this limitation, we remove the reconstruction term from these objectives and incorporate negative guidance to design a \textbf{Controllable Stylized Distillation (CSD)}, which leverages 2D stylized priors to facilitate faithful and structurally coherent style transfer in 3D scenes.

Unlike existing approaches that primarily rely on VGG-based features, to the best of our knowledge, our method is the first 3DGS-based style transfer framework that relies solely on diffusion model distillation. It enables flexible extension of 2D style transfer techniques to 3D scenes, bridging the gap between 2D and 3D stylization.
We evaluate our method through both qualitative and quantitative comparisons against baseline approaches, demonstrating superior performance in style transfer quality and content preservation. In summary, our key contributions are as follows:
\begin{itemize}
    \item We propose a Multi-View Frequency Consistency that enhances cross-view consistency by preserving high-frequency components and selectively suppressing low-frequency components.
    \item We propose a Controllable Stylized Distillation that prevents content leakage from 2D stylized priors and optimizes 3D scenes. 
    \item Extensive experiments demonstrate that our method outperforms existing state-of-the-art approaches in both qualitative and quantitative evaluations.
    \item To the best of our knowledge, FantasyStyle is the first 3DGS-based style transfer method that relies entirely on diffusion model distillation. It enables the flexible extension of 2D style transfer techniques to 3D domains.
\end{itemize}
\section{Related Work}
\label{sec:related}
\begin{figure*}[h!]
    \centering
    \includegraphics[width=\linewidth]{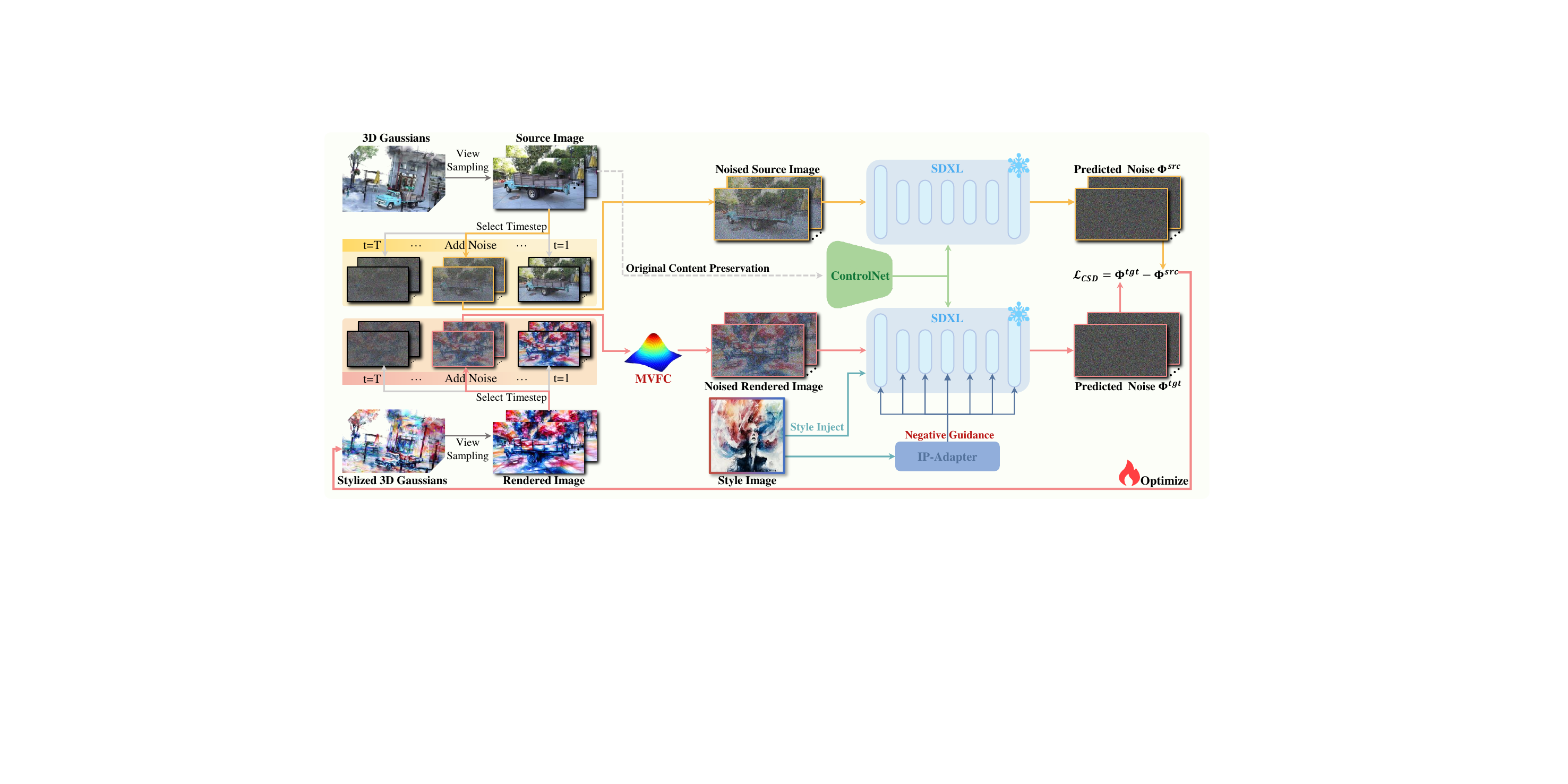}
    \caption{Overview of our proposed method. Inspired by DDS~\cite{hertz2023delta},  FantasyStyle consists of two distinct pathways: Source Image and Rendered Image, each highlighted in different colors. We propose Controllable Stylized Distillation (CSD) to optimize the 3D scene. In Rendered Image pathway, we introduce a multi-view frequency consistency (MVFC) and inject style image features to obtain a multi-view consistent 2D stylized prior. Additionally, we incorporate negative guidance to suppress potential content leakage from the style image.}
    \label{fig:model}
\end{figure*}

\noindent \textbf{2D Style Transfer Based on Diffusion Models}. Style transfer aims to separate style and content representations from a style image and apply the style to another image. With the rise of diffusion models, 2D style transfer has seen notable advances. Recent studies leverage cross-attention~\cite{zhang2023inversion} and self-attention mechanisms~\cite{hertz2024style,chung2024style} to inject style features into the generation process. In addition, frequency-domain techniques~\cite{xu2025stylessp,gao2024frequency} have been incorporated into the diffusion process, effectively enhancing the model’s ability to control stylistic attributes. More recently, methods~\cite{frenkel2024implicit,ouyang2025k} based on Low-Rank Adaptation (LoRA)~\cite{hu2022lora} have been proposed to achieve a more effective separation of content and style features, thereby improving the flexibility and diversity of style transfer. We propose a Controllable Stylized Distillation approach that utilizes 2D stylized priors to optimize 3D scenes, effectively bridging the gap between 2D and 3D style transfer.


\noindent \textbf{3DGS Style Transfer Based on VGG}. 3D style transfer is a conditional generative task that applies the visual style of a 2D image to a 3D representation while preserving the original geometric structure. The VGG network, one of the earliest architectures used in style transfer, remains a widely adopted feature extractor in both 2D and 3D tasks. Saroha et al.\cite{saroha2024gaussiansplattingstyle} employ VGG and AdaIN to transfer styles to specific views. StyleSplat\cite{StyleSplat} introduces a nearest neighbor feature matching loss between VGG features of rendered and reference images. SGSST~\cite{galerne2024sgsst} proposes a Simultaneously Optimized Scales loss to enable ultra-high-resolution 3D stylization. StyleGaussian~\cite{StyleGaussian} embeds 2D VGG scene features into transformed 3D Gaussian features, which are fused with reference features and decoded into stylized RGB outputs for more coherent, view-consistent stylization. However, these methods often suffer from content leakage from the style image. To address this, we propose a negative guidance mechanism to suppress the transfer of irrelevant content information from the style image.

\noindent \textbf{3DGS Style Transfer Based on Diffusion Model}. With the powerful generative capacity of text-to-image (T2I) diffusion models, recent 3D style transfer methods increasingly incorporate pre-trained diffusion priors to guide stylization. InstantStyleGaussian~\cite{InstantStyleGaussian} accelerates the process by generating target-style images via diffusion and integrating them into the training set to iteratively optimize the 3D Gaussian Splatting scene. ArtNVG~\cite{ArtNVG} enhances content-style disentanglement using CSGO~\cite{xing2024csgo} and Tile-ControlNet~\cite{zhang2023adding}, and introduces an attention-based view alignment module for consistent local texture and color. StyleMe3D~\cite{zhuang2025styleme3d} proposes dynamic style score distillation in the latent space of Stable Diffusion for improved semantic alignment. These methods, when guided by 2D images for 3D scene stylization, do not explicitly enforce multi-view consistency. To address this, we propose a Multi-View Frequency Consistency to improve consistency across views.
\section{Preliminary}
\noindent \textbf{Score Distillation Sampling (SDS)}~\cite{poole2022dreamfusion} is widely used in 3D generation, leveraging pretrained diffusion models with rich 2D priors to distill knowledge into 3DGS models. Ignoring the UNet Jacobian, it is formulated as:
\begin{equation}
\nabla_{\theta}\mathcal{L}_{\mathrm{SDS}}=\mathbb{E}_{t,\epsilon}\left[\omega(t)\left(\epsilon_{\phi}(z_t\mid \mathcal{P})-\epsilon\right)\frac{\partial z_t}{\partial\theta}\right],
\label{eq:sds}
\end{equation}
where $\omega(t)$ denotes a time-dependent weighting function, $\epsilon\sim\mathcal{N}(0,\textbf{I})$, $z_t = \sqrt{\overline{\alpha}_t} \mathbf{x} + \sqrt{1 - \overline{\alpha}_t} \epsilon$ represents the noised input at timestep $t$, and $\epsilon_{\phi}(z_t \mid y)$ is the predicted denoising score conditioned on the text prompt $\mathcal{P}$.

\noindent \textbf{Delta Denoising Score (DDS)}~\cite{hertz2023delta} extends SDS for editing tasks by estimating the score difference between target and source distributions to guide updates. This mitigates the noisy gradients of SDS that often lead to blurred or unclear images. It is defined as:
\begin{equation}
\begin{aligned}
\nabla_\theta\mathcal{L}_{\mathrm{DDS}} = \mathbb{E}_{t,\boldsymbol{\epsilon}}\Big[
  \omega(t)\big(& \boldsymbol{\epsilon}_\phi(z_t^\mathrm{tgt}, \mathcal{P}^\mathrm{tgt}, t) \\
  & - \boldsymbol{\epsilon}_\phi(z_t^\mathrm{src}, \mathcal{P}^\mathrm{src}, t) \big)
  \frac{\partial z_t^\mathrm{tgt}}{\partial\theta}
\Big],
\label{eq:dds}
\end{aligned}
\end{equation}
where $z_t^{\mathrm{tgt}}$ and $z_t^{\mathrm{src}}$ denote the latent representations at timestep $t$, sharing the same noise $\epsilon_t$.

\noindent \textbf{Task Definition.}  The 3DGS reconstruction process can be formally defined as follows:
\begin{equation}
\min_\Theta\frac{1}{N}\sum_{i=1}^N \mathcal{L}_{L1+SSIM}(\mathcal{R}(C_i;\Theta),V_i^{gt}),
\label{eq:gs}
\end{equation}
where $C_i$ denotes the $i$-th camera, and $\Theta = \{(u_{j}, \Sigma_{j}, \alpha_{j}, c_{j,0}, (c_{j,k,l})_{k,l})\}^{M}_{j=1}$
represents the set of 3D Gaussian scene parameters. $c_{i,0}$ corresponds to the spherical harmonics (SH) basis functions, while $c_{j,k,l}$ are the SH coefficients that control the view-dependent color appearance. $\mathcal{R}(\cdot)$ denotes the rasterization-based rendering process that produces the corresponding view, and $V^{gt}_{i}$ refers to the ground truth image for camera $C_i$.

The objective of style transfer is to modify the appearance of a scene without altering its underlying geometry. Accordingly, we reformulate Eq.~\ref{eq:gs} by optimizing only the color-related terms while keeping other parameters fixed. This can be formally expressed as:
\begin{equation}
\min_{\Theta_{c}}\frac{1}{N}\sum_{i=1}^N\mathcal{L}(\mathcal{R}(C_i;\Theta_{c}); P,I_{style}),
\end{equation}
where $P$ denotes the prompt describing the original scene, and $I_{style}$ represents the reference style image.

\begin{figure}[]
    \centering
    \includegraphics[width=\linewidth]{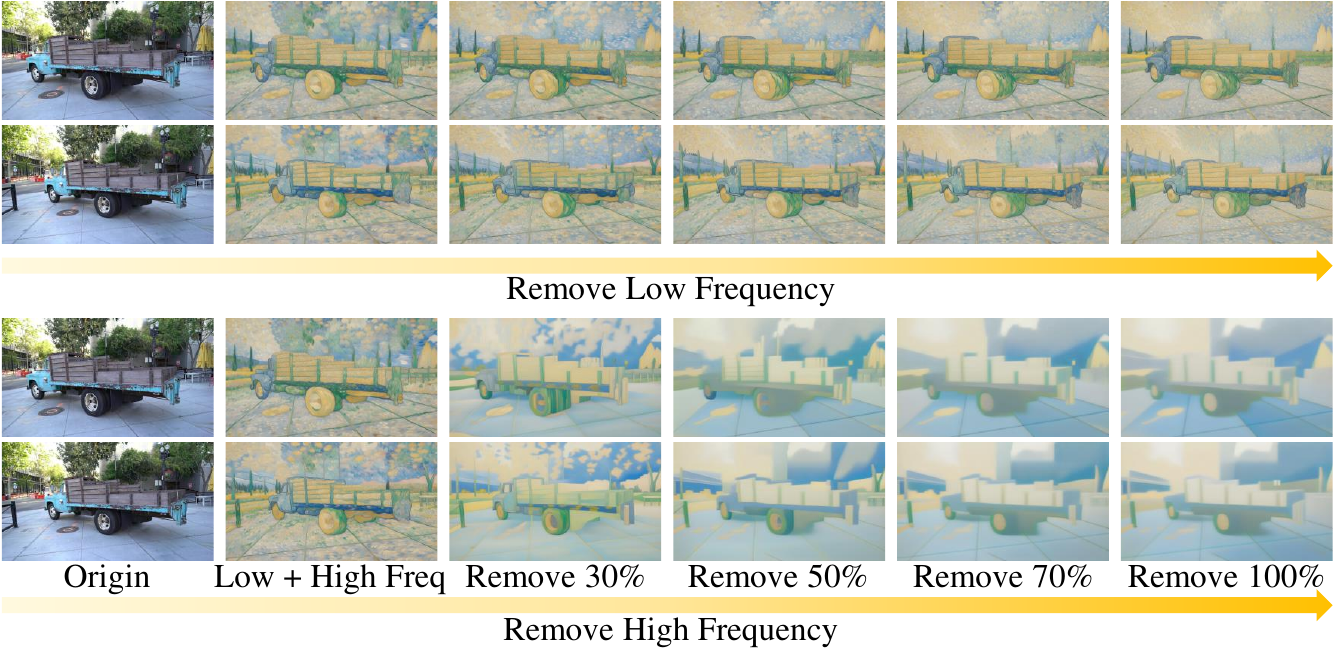}
    \caption{The role of different frequency components. For intuitive visualization, we present multi-view 2D stylization results. We observe that selectively removing low-frequency components slightly reduces local detail while significantly improving multi-view consistency, whereas removing high-frequency components severely degrades texture features, resulting in blurred appearances.}
    \label{fig:freq}
\end{figure}
\section{Method}
We present the proposed FantasyStyle framework in detail (see Fig.~\ref{fig:model}). To address the issue of multi-view inconsistency, we apply a Multi-View Frequency Consistency~(Sec.~\ref{sec:mvfc}) on the DDIM-noised multi-view latent, which enhances view consistency and mitigates conflicts arising from inconsistent stylized priors. Furthermore, we introduce a Controllable Stylized Distillation~(Sec.~\ref{sec:csd}) that suppresses content information from the style image and leverages 2D stylized priors to optimize the 3D scene, enabling effective 3D style transfer.
\subsection{Multi-View Frequency Consistency}
\label{sec:mvfc}
\noindent \textbf{3D Frequency Analysis}. Inspired by FreeU~\cite{si2024freeu} and FreeInit~\cite{wu2024freeinit}, which emphasize the critical role of frequency components in image and video generation, we perform a frequency-domain analysis of the multi-view stylized priors to better understand their behavior. To visualize more intuitively, we use IP-Adapter~\cite{ye2023ip} to generate 2D multi-view stylized images. Moreover, since our task only involves fine-tuning color parameters to achieve texture-level style consistency without the need to preserve geometric structure, we introduce ControlNet to help guide the structural information in the images.
Given $N$ rendered views from a 3D scene, we first apply DDIM~\cite{ho2020denoising} to add noise and obtain $z_t$:
\begin{equation}
z_t^{N}=\sqrt{\overline{\alpha}_t}z_0^{N}+\sqrt{1-\overline{\alpha}_t}\epsilon^N,
\end{equation}
where $\overline{\alpha}_t$ is a constant, and $\epsilon^N \sim\mathcal{N}(0,\textbf{I})$ represents Gaussian white noise. We investigate the impact of different frequency components on style consistency across 3D views by decomposing $z_t$ into its low-frequency and high-frequency components, as formulated below:
\begin{equation}
\begin{aligned}
F_{z_{t}}^{L} &= FFT_{3D}(z_{t}^N) \odot (1-\mathcal{H}), \\
F_{z_{t}}^{H} &= FFT_{3D}(z_{t}^N) \odot \mathcal{H}, \\
{z'}_{t}^{N} &= IFFT_{3D}(\alpha * F_{z_{t}}^{L} + F_{z_{t}}^{H}), &where\: \alpha \in [0,1]\\
{z''}_{t}^{N} &= IFFT_{3D}(F_{z_{t}}^{L} + \alpha *F_{z_{t}}^{H}), & where\: \alpha \in [0,1]
\end{aligned}
\end{equation}
where $FFT_{3D}$ denotes the Fast Fourier Transform applied to both multi-view spatial and batch dimensions, $IFFT_{3D}$ is its inverse, and $\mathcal{H}$ is a High Pass Filter (e.g., Gaussian). $\odot$ denotes element-wise multiplication. We control the parameter $\alpha$ to selectively remove specific frequency components, and the corresponding results are shown in Fig.~\ref{fig:freq}. We observe that removing low-frequency components does not significantly affect stylization performance. Moderately reducing low-frequency components slightly diminishes local details but significantly improves multi-view consistency. However, when low-frequency components are overly suppressed, the image loses most local details and appears noticeably brighter.
In contrast, as high-frequency components are progressively removed, the images lose fine-grained texture details and become overly smooth. 
\begin{figure}[]
    \centering
    \includegraphics[width=\linewidth]{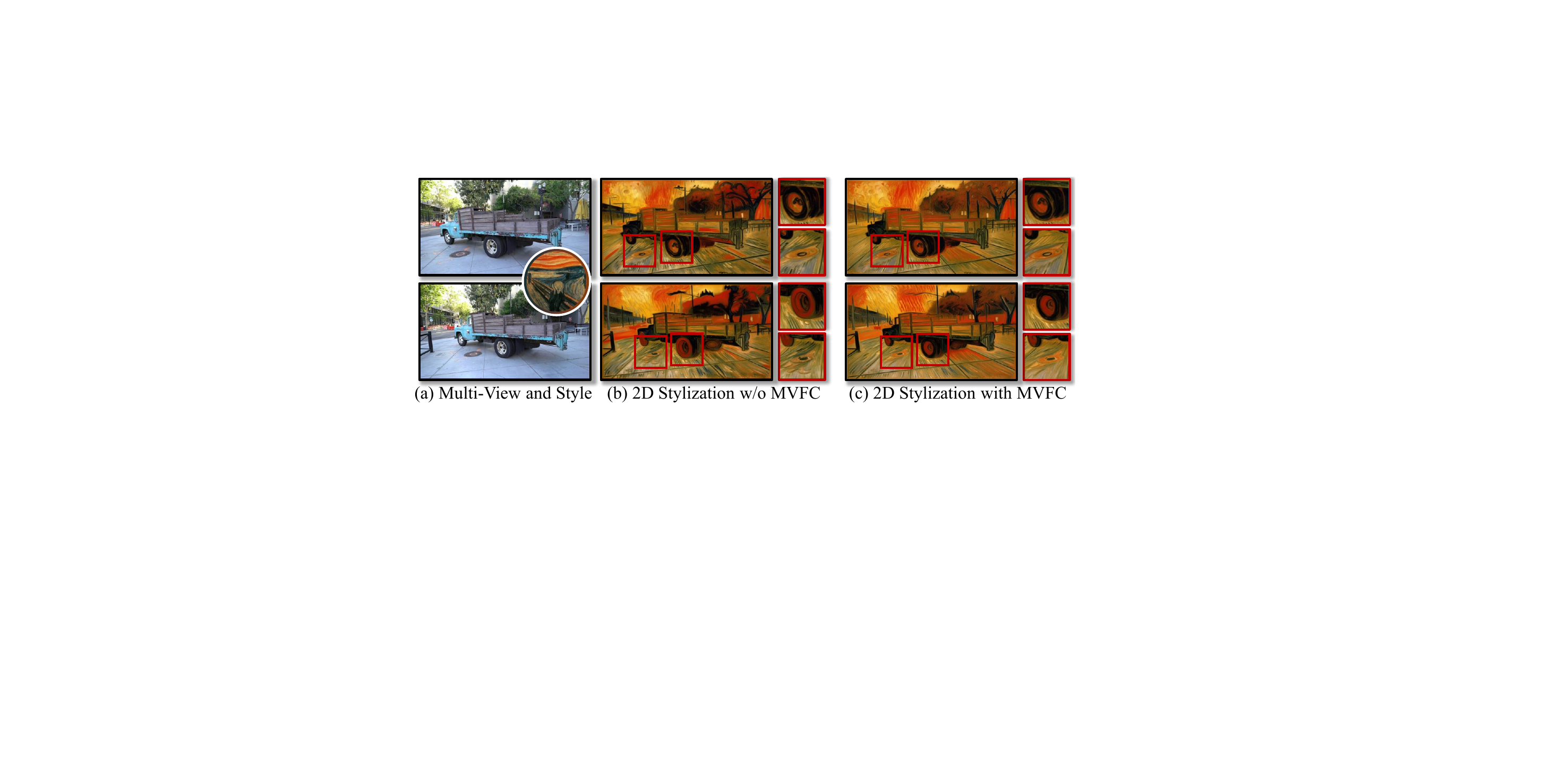}
    \caption{Visual results of MVFC on multi-view 2D stylization. Without MVFC, inconsistent stylization conflicts arise across views.}
    \label{fig:multi-view}
\end{figure}

\noindent \textbf{Multi-View Frequency Control}. Based on the above observations, we propose a Multi-View Frequency Consistency (MVFC) mechanism to address view inconsistency in 3D style transfer. Specifically, we retain all high-frequency components to preserve texture details, while selectively preserving the low-frequency components to balance local details and multi-view consistency. Additionally, we introduce a shared low-frequency component of random Gaussian noise across all views to explicitly enforce consistency. The mathematical formulation is as follows:
\begin{equation}
\begin{aligned}
F_{\epsilon^{\prime}}^{L} &= FFT_{3D}(\epsilon^{\prime}) \odot (1-\mathcal{H}),\quad \quad \epsilon^{\prime} \sim\mathcal{N}(0,\textbf{I})\\
\hat{z}_{t}^{N} &= IFFT_{3D}(\gamma * F_{z_{t}}^{L} + (1-\gamma)*F_{\epsilon^{\prime}}^{L} +F_{z_{t}}^{H}),
\end{aligned}
\end{equation}
where $\gamma$ is a balancing factor. We use 2D stylized images to visually demonstrate the issue of view inconsistency, as shown in Fig.~\ref{fig:multi-view}. Without MVFC, significant style conflicts occur across different views. In contrast, incorporating MVFC substantially improves multi-view style consistency, enabling us to obtain coherent 2D stylized priors that can guide the stylization of 3D scenes through subsequent Controllable Stylized Distillation.

\subsection{Controllable Stylized Distillation}
\label{sec:csd}
\noindent \textbf{Style Injection}. As shown in Fig.~\ref{fig:model}, we build upon Stable Diffusion XL (SDXL) and incorporate IP-Adapter for style injection, aiming to obtain 2D stylized priors. Although the 3D scene's geometry remains fixed during optimization, generating 2D stylized priors often causes geometric information loss, leading to structural inconsistencies across views. To address this, we incorporate ControlNet~\cite{zhang2023adding} to explicitly guide structural features during stylization, improving the geometric consistency of the 2D stylized priors.

\noindent \textbf{Negative Guidance}. The 2D stylized priors obtained through the above method, similar to previous approaches based on VGG features, often suffer from inevitable content leakage from the style image, leading to suboptimal performance in 3D style transfer. Inspired by recent advances in 2D image generation, we introduce negative prompts into the Classifier-Free Guidance (CFG) mechanism to suppress content interference from the style image. The formulation of the classifier-free guidance is as follows:
\begin{equation}
\tilde{\epsilon}_\phi(z_t,t,\mathcal{P},\varnothing)=\epsilon_\phi(z_t,t,\varnothing)+ 
\beta\left(\epsilon_\phi(z_t,t,\mathcal{P})-\epsilon_\phi(z_t,t,\varnothing)\right),
\label{eq:cfg}
\end{equation}
where $\mathcal{P}$ denotes the text prompt, $\varnothing$ represents the null-text embedding, and $\beta$ is the CFG scale. To suppress specific content information, $\varnothing$ is replaced with the negative prompt $\mathcal{P}_{neg}$, thereby enabling negative guidance. Negative prompts can be generated using large multimodal language models. For convenience, we leverage the IPAdapter-Instruct (IP)~\cite{rowles2024ipadapter} to extract both style and content features from the style image, which are then used to construct the negative prompt. Accordingly, Eq.~\ref{eq:cfg} is reformulated as follows:
\begin{equation}
\begin{aligned}
\hat{\epsilon}_\phi(z_t,t, &\mathcal{P}, \text{IP}(I_{r})^{s},\text{IP}(I_{r})^{c}) =\epsilon_\phi(z_t,t,\text{IP}(I_{r})^{c})+ \\
&\beta\left(\epsilon_\phi(z_t,t,[\mathcal{P},\text{IP}(I_{r})^{s}])-\epsilon_\phi(z_t,t,\text{IP}(I_{r})^{c})\right),
\label{eq:neg}
\end{aligned}
\end{equation}
where $\text{IP}(I_{r})^{s}$ and $\text{IP}(I_{r})^{c}$ denote the content and style features extracted from the reference style image, respectively, and $[,]$ represents the concatenation operation. 

\noindent \textbf{Controllable Stylized Distillation}. 
Directly applying SDS or DDS to optimize 3D scenes using 2D stylized priors results in overly smooth and blurry images, losing crucial brushstroke details essential for style transfer, as shown in Fig.~\ref{fig:ablation2}. To better understand this limitation, we define $\delta_{z_t}^{\mathrm{SDS}}:=\boldsymbol{\epsilon}_\phi(z_t,t,\mathcal{P})-\boldsymbol{\epsilon}.$ in Eq.~\ref{eq:sds} and $\delta_{z_t}^{\mathrm{DDS}}:=\boldsymbol{\epsilon}_\phi(z_t^\mathrm{tgt}, \mathcal{P}^\mathrm{tgt}, t)- \boldsymbol{\epsilon}_\phi(z_t^\mathrm{src}, \mathcal{P}^\mathrm{src}, t)$ in Eq.~\ref{eq:dds}, and extend it based on Eq.~\ref{eq:cfg} as follows:
\begin{equation}
\begin{aligned}
&\delta_{z_t}^\mathrm{SDS}:=\underbrace{\epsilon_\phi(z_t,t,\varnothing)-\boldsymbol{\epsilon}}_{\delta_{z_t}^\mathrm{recon}}+\beta\underbrace{(\boldsymbol{\epsilon}_\phi(z_t,t,\mathcal{P})-\boldsymbol{\epsilon}_\phi(z_t,t,\varnothing))}_{\delta_{z_t}^\mathrm{cfg}},
\\
&\delta_{z_t}^\mathrm{DDS}:=\underbrace{\boldsymbol{\epsilon}_\phi(z_t^\mathrm{tgt},t,\varnothing)-\boldsymbol{\epsilon}_\phi(z_t^\mathrm{src},t,\varnothing)}_{\delta_{z_t}^\mathrm{recon}}+\beta(\delta_{z_t^\mathrm{tgt}}^\mathrm{cfg}-\delta_{z_t^\mathrm{src}}^\mathrm{cfg}).
\end{aligned}
\label{eq:dds_simple}
\end{equation}

The two formulations above are highly similar and can be decoupled into a reconstruction term $\delta_{z_t}^\mathrm{recon}$ and a classifier-free guidance term $\delta_{z_t}^\mathrm{cfg}$. Since our task involves only fine-tuning color-related parameters without considering geometric structure or identity preservation, the reconstruction term becomes a limiting factor. It not only leads to overly smooth and blurry results but also significantly slows down the optimization process. To address this issue, we remove the reconstruction term and introduce negative guidance. Building upon Eq.~\ref{eq:dds}, Eq.~\ref{eq:neg}, and Eq.~\ref{eq:dds_simple}, we propose Controllable Stylized Distillation (CSD), formulated as follows:
\begin{align}
&\Phi^{tgt} = \beta(\epsilon_\phi(z_t^\mathrm{tgt},t,[\mathcal{P},\text{IP}(I_{r})^{s}])-\epsilon_\phi(z_t^\mathrm{tgt},t,\text{IP}(I_{r})^{c})), \notag \\
&\Phi^{src} = \beta(\epsilon_\phi(z_t^\mathrm{src},t,\mathcal{P})-\epsilon_\phi(z_t^\mathrm{src},t,\varnothing)), \label{eq:csd} \\
&\nabla_\theta\mathcal{L}_{\mathrm{CSD}} = \mathbb{E}_{t,\boldsymbol{\epsilon}}[
  \omega(t)(\Phi^{tgt}-\Phi^{src}) \frac{\partial z_t^\mathrm{tgt}}{\partial\theta}
]. \notag
\end{align}

Unlike SDS and DDS, which randomly sample from the entire diffusion timestep range, we perform random sampling from a fixed set of discrete timesteps to better simulate the DDIM denoising process of the diffusion model. As illustrated in Fig.~\ref{fig:model}, a timestep 
$t$ is randomly selected to perturb both the source and rendered images, after which Eq.~\ref{eq:csd} is computed for each perturbed pair to optimize the 3D Gaussian scene. FantasyStyle is detailed in Alg.~\ref{alg:1}.
\begin{algorithm}[!t]
\caption{FantasyStyle}
\begin{algorithmic}[1]
\State \textbf{Input:} Diffusion model $\epsilon_\phi$, iterations $M$, guidance scale $\beta$ ,origin text embedding $\mathcal{P}$, content and style embeddings of the style image $\text{IP}(I_{r})^{c}$ and $\text{IP}(I_{r})^{s}$.
\State \textbf{Output:} 3D Stylized Scene.
\For{$m = 1, 2, \ldots, M$}
    \State Sample: $z_{0}^{N} = \mathcal{R}(C_N;\Theta)$, $\epsilon^N \sim\mathcal{N}(0,\textbf{I})$, random timestep $t$, view number $N$
    \For{$k = [src,tgt],P_{pos} = [\mathcal{P},[\mathcal{P},\text{IP}(I_{r})^{s}]],P_{neg}=[\varnothing,\text{IP}(I_{r})^{c}]$}
        \State $z_t^{N}=\sqrt{\overline{\alpha}_t}z_0^{N}+\sqrt{1-\overline{\alpha}_t}\epsilon^N$
        \If{k=tgt} 
            \State $\hat{z}_{t}^{N}  = \text{MVFC}(z_t^{N})$
        \EndIf
        \State Predict $\epsilon_\phi(z_t^{N},t,P_{pos})$ and $\epsilon_\phi(z_t^{N},t,P_{neg})$
        \State $\Phi = \beta(\epsilon_\phi(z_t^{N},t,P_{pos})- \epsilon_\phi(z_t^{N},t,P_{neg}))$
    \EndFor
    \State $\nabla_\theta\mathcal{L}_{\mathrm{CSD}} =\omega(t)(\Phi^{tgt}-\Phi^{src})$ Eq.~(\ref{eq:csd})
    \State Update ${z_{0}^{tgt}}^{N}$ with $\nabla_\theta\mathcal{L}_{\mathrm{CSD}}$
    \State Update 3D Gaussians
\EndFor
\State \textbf{Return:} 3D Stylized Scene.
\end{algorithmic}
\label{alg:1}
\end{algorithm}
\section{Experiments}
\begin{figure*}[]
    \centering
    \includegraphics[width=\linewidth]{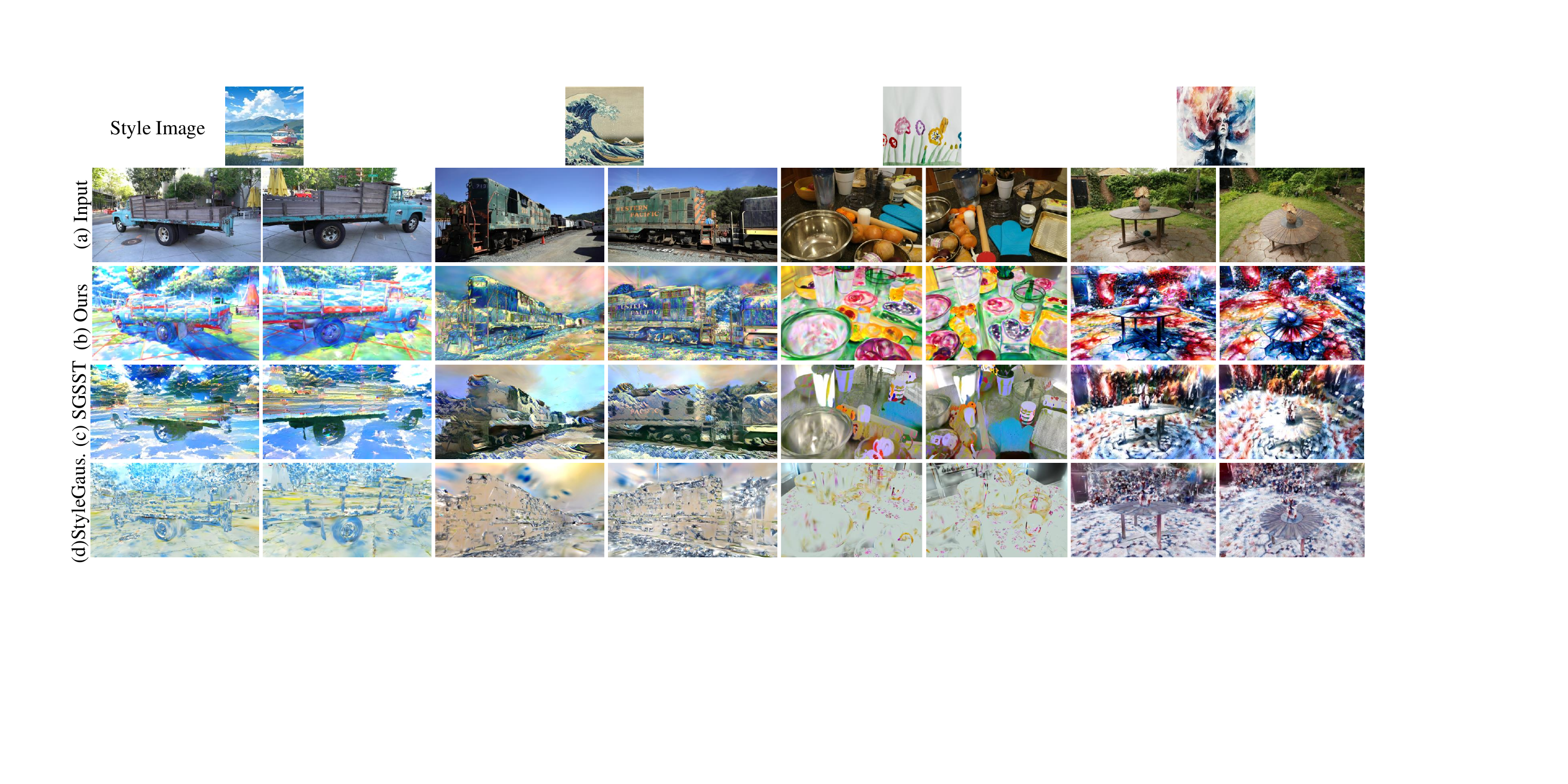}
    \caption{Qualitative comparison of different methods. Our approach achieves superior style transfer quality compared to existing methods. Zoom in for better view.}
    \label{fig:comparison}
\end{figure*}
\begin{table*}[!t]
\centering
\begin{tabular}{cccccccc}
\hline
\multirow{2}{*}{Method} & \multirow{2}{*}{ArtFID $\downarrow$} & \multirow{2}{*}{$\text{FID}_{style} \downarrow$}  & \multirow{2}{*}{$\text{FID}_{content}$ $\downarrow$}            & \multicolumn{2}{c}{Short-range consistency} & \multicolumn{2}{c}{Long-range consistency} \\ \cline{5-8} 
                        &              &           &                                   & LPIPS  $\downarrow$              & RMSE  $\downarrow$               & LPIPS   $\downarrow$             & RMSE  $\downarrow$              \\ \hline
StyleGaussian           & 45.31    &  398.17 & 331.53                &\underline{ 0.290 }            & \underline{0.267 }            & \underline{0.542}              & \textbf{0.425}            \\
SGSST                   &\underline{ 44.70 }  & \underline{370.03}   & \underline{314.09  }              & 0.295            & 0.27           & 0.569             & 0.502            \\ \hline
Ours                     & \textbf{43.52}   & \textbf{347.61 } & \textbf{261.71} & \textbf{0.285  }           & \textbf{0.241  }           & \textbf{0.529  }           & \underline{0.462}           \\ \hline
\end{tabular}
\caption{Quantitative comparison of different methods in 3DGS style transfer. $\text{FID}_{style}$ measures the style fidelity to style image, while $\text{FID}_{content}$ evaluates content preservation to the original input. \textbf{Bold}: best; \underline{underline}: second best.}
\label{tab:comparison}
\end{table*}
\subsection{Experimental Setup}
\noindent \textbf{Datasets}. Our evaluation utilizes several distinct scene datasets (IN2N~\cite{haque2023instruct}, Tandt DB~\cite{kerbl20233d}, and Mip-NeRF 360~\cite{barron2022mip}), with 8 different image styles applied to each scene.

\noindent \textbf{Metrics}. We use Art-FID~\cite{wright2022artfid} to jointly evaluate content and style preservation. LPIPS~\cite{zhang2018unreasonable} and RMSE measure short-term and long-term content consistency with respect to the input, while FID assesses both style fidelity to the style image and content preservation relative to the original input.


\noindent \textbf{Comparison Methods}. We compare our method with recent state-of-the-art 3DGS-based style transfer methods, including SGSST~\cite{galerne2024sgsst} and StyleGaussian~\cite{StyleGaussian}. 

\noindent \textbf{Implementation Detail}. We deploy 3DGS, SDXL~\cite{podell2023sdxl}, ControlNet~\cite{zhang2023adding}, and IP-Adapter~\cite{ye2023ip} in all experiments. DDIM is used for noise sampling, with the CFG scale $\beta$ set to 7.5. For MVFC, the parameter $\gamma$ is set to 0.9. All experiments are conducted on 2 NVIDIA L20 (48GB) GPUs.

\begin{figure}[!t]
    \centering
    \includegraphics[width=\linewidth]{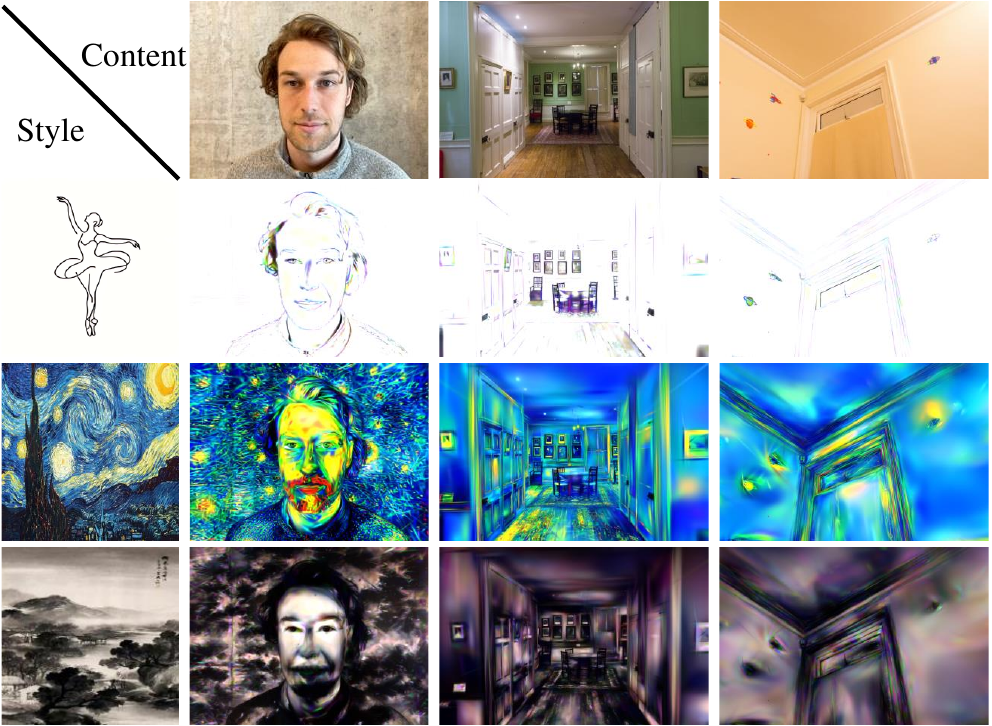}
    \caption{More qualitative results from our method.}
    \label{fig:show}
\end{figure}
\begin{figure}[!t]
    \centering
    \includegraphics[width=\linewidth]{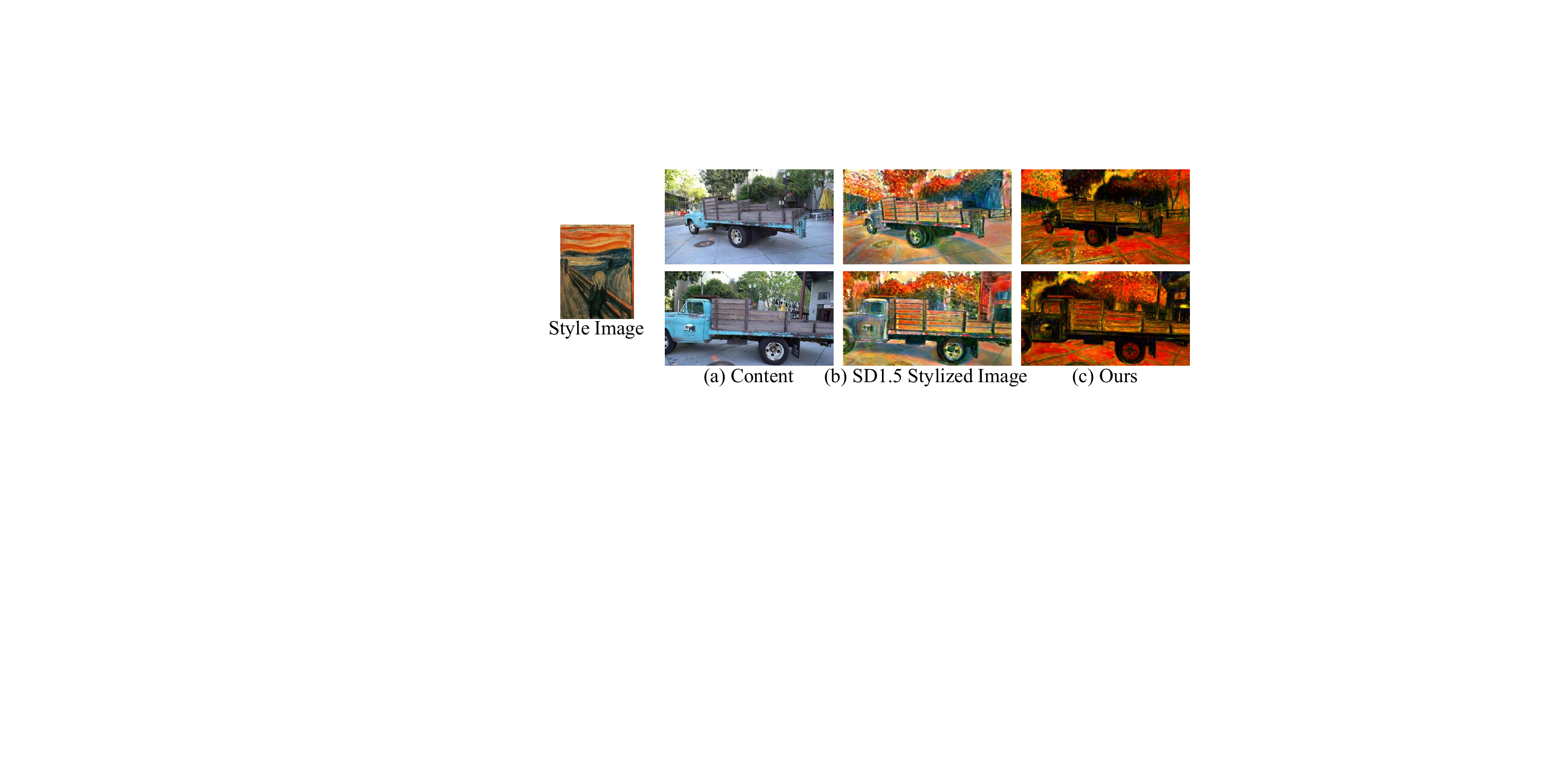}
    \caption{Adaption to other 2D style transfer methods.}
    \label{fig:others}
\end{figure}

\subsection{Comparison Results}

\noindent \textbf{Quantitative Comparisons}. To comprehensively evaluate the performance of our method, we conduct systematic comparisons with existing approaches across multiple metrics, as shown in Tab.~\ref{tab:comparison}. Our method consistently outperforms SGSST and StyleGaussian on key metrics such as ArtFID and FID, demonstrating superior style transfer quality and content preservation. Although prior methods are often optimized using VGG-based perceptual features, which may favor LPIPS scores, our approach still achieves the best results in both short-term and long-term LPIPS evaluations. These results confirm the effectiveness of our method in maintaining content and style consistency across views.

\noindent \textbf{Qualitative Comparisons}. Fig.~\ref{fig:comparison} presents qualitative comparisons across four different scenes and various styles. Although StyleGaussian achieves LPIPS scores comparable to ours, its visual quality is significantly inferior. It fails to effectively transfer the target style’s color and brushstroke characteristics and often introduces artifacts that compromise the integrity of the original content. In contrast, SGSST demonstrates better style transfer in terms of color and stroke patterns; however, it tends to over-stylize the results. As seen in the first, second, seventh, and eighth columns of the third row, it introduces excessive stylistic elements, such as dense brushstrokes, which overwhelm and distort the original structure. Although these methods only optimize color parameters, they still disrupt the geometry of the rendered images. Moreover, due to the inherent limitations of VGG-based features, existing methods often focus excessively on the appearance of the style image, rather than extracting abstract, transferable style representations adaptable to diverse input content. In comparison, our approach benefits from diffusion-based priors, enabling it to extract high-level, abstract style features and apply them consistently across varied scenes, while preserving the original structure of the image. We also present additional results, as shown in Fig.~\ref{fig:show}. Overall, our method achieves a better visual balance between stylization quality and content preservation, offering both faithful style transfer and structural integrity.

\noindent \textbf{Other Extensions}. As shown in Fig.~\ref{fig:others}, our method can be easily extended to incorporate various 2D style transfer models. Enhancements in 2D stylization directly translate to improved 3D visual quality.


\noindent \textbf{Discussion}. While VGG-based methods have fallen out of favor in 2D style transfer due to their limited ability to capture high-level style semantics and prevent content leakage, most 3DGS-based approaches still rely on them, inheriting these limitations. In contrast, recent 2D methods have shifted toward diffusion-based techniques for their flexibility and expressive power. However, 3DGS-based style transfer frameworks distilled purely from diffusion models remain unexplored. Our method introduces a novel pipeline that bridges this gap, enabling the extension of diffusion-based 2D style transfer techniques to 3D scenes.


\begin{figure}[]
    \centering
    \includegraphics[width=\linewidth]{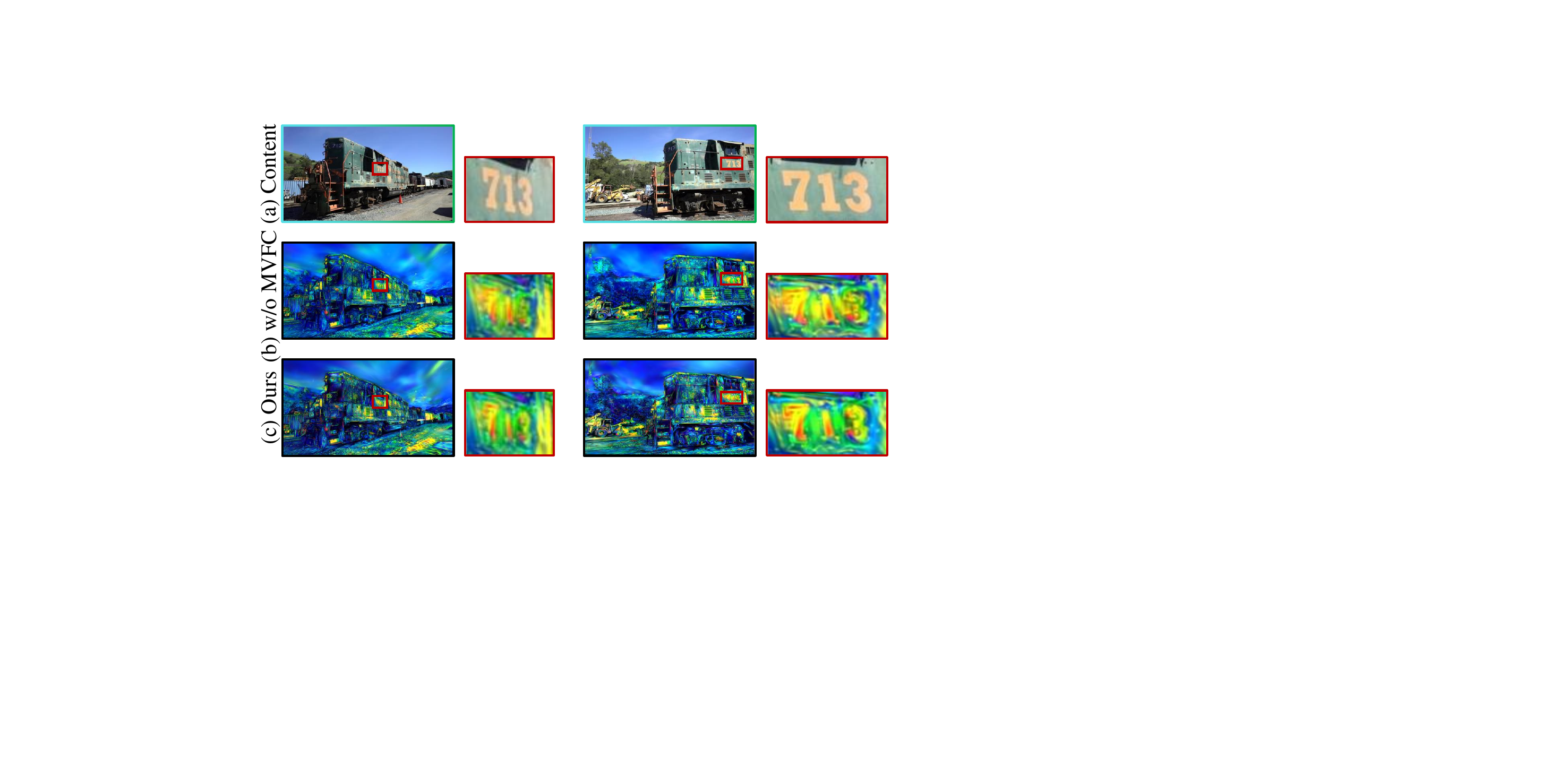}
    \caption{Ablation visual results on MVFC.}
    \label{fig:ablation1}
\end{figure}
\subsection{Ablation Study}
\begin{table}[]
\centering
\small
\setlength{\tabcolsep}{4pt} 
\begin{tabular}{ccccc}
\hline
\multirow{2}{*}{Method}   & \multicolumn{1}{c}{Short-range consistency} & \multicolumn{1}{c}{Long-range consistency} \\
                                & LPIPS  $\downarrow$               & LPIPS   $\downarrow$            \\ \hline

w/o MVFC              & 0.253            & 0.587                \\

Ours                    & \textbf{0.250 }                  & \textbf{0.574  }                  \\ \hline
\end{tabular}
\caption{Quantitative results of the ablation study on MVFC.}
\label{tab:mvfc}
\vspace{-10pt}
\end{table}

\noindent \textbf{Ablation Study on Multi-View Frequency Consistency}. To thoroughly validate the effectiveness of MVFC, we conduct both quantitative and qualitative experiments. As shown in Tab.~\ref{tab:mvfc}, our method achieves consistent improvements on LPIPS metrics under both short-term and long-term consistency evaluations. Notably, since MVFC is specifically designed to address multi-view inconsistency, the gains are more pronounced in the long-term LPIPS results. Qualitative comparisons in Fig.~\ref{fig:ablation1} further support these findings. Without MVFC, conflicting 2D stylized priors often lead to blurred edge details and structural distortion in the reconstructed content. In contrast, MVFC effectively mitigates these conflicts, preserving sharp object boundaries and maintaining structural integrity across views.

\begin{figure}[!t]
    \centering
    \includegraphics[width=\linewidth]{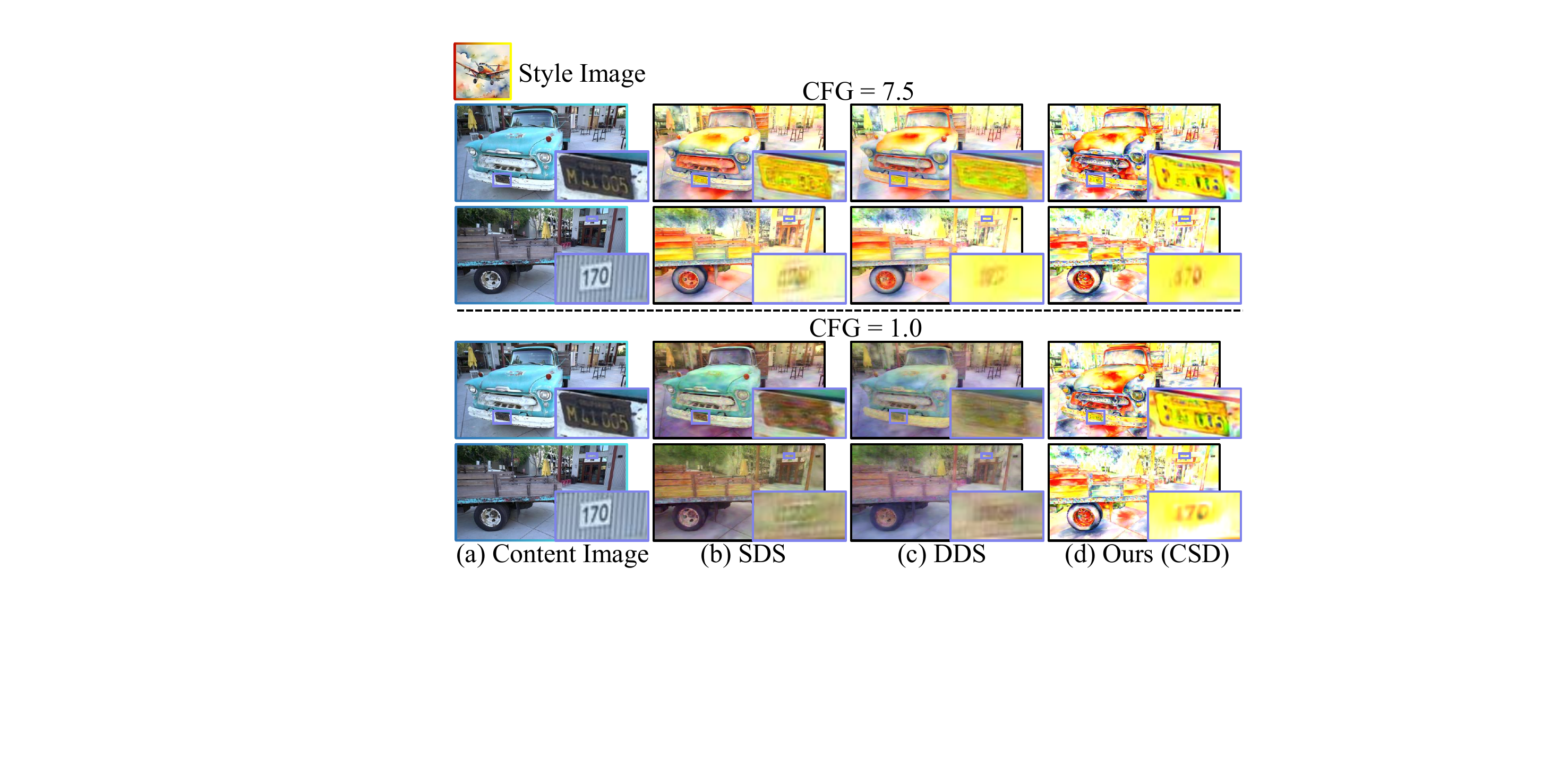}
    \caption{Ablation results on optimization strategy and CFG.}
    \label{fig:ablation2}
\end{figure}
\noindent \textbf{Ablation Study on Optimization Strategy.} Due to the presence of the reconstruction term, SDS and DDS often produce overly smooth and blurry outputs. In style transfer tasks, this is undesirable, as it leads to the loss of essential stylistic features, particularly brushstroke details. As illustrated in Fig.~\ref{fig:ablation2}, while SDS and DDS successfully transfer the color characteristics of the style image, they fail to preserve brushstroke textures and significantly blur the structural details of the original content. Moreover, other optimization strategies require careful CFG scale tuning across styles or scenes, increasing complexity. In contrast, our method removes the reconstruction term, making the CFG scale less sensitive. As a result, adjusting the CFG scale does not affect stylistic expressiveness.

\subsection{Limitations}
Although SDXL-based optimization is time-consuming, it can be mitigated by using smaller models, lower resolutions, or adjusted color-related learning rates.

\section{Conclusion}
We propose FantasyStyle, a 3DGS-based style transfer framework that fully exploits diffusion model distillation. It features two key components: (1) Multi-View Frequency Consistency, which applies a 3D frequency filter to DDIM-noised multi-view latents to preserve high-frequency details and suppress low-frequency components for better view consistency; and (2) Controllable Stylized Distillation, which introduces negative guidance to prevent content leakage and uses 2D stylized priors to optimize the 3D scene. Experiments show that our method outperforms existing approaches in visual quality and content preservation.


\section*{Acknowledgments}
This project is sponsored by Shanghai Pujiang Programme 24PJD030 and Natural Science Foundation of Shanghai 25ZR1402138. Changshuo Wang is supported by the European Union’s Horizon 2024 Research and Innovation Programme for the Marie Skłodowska-Curie Actions under Grant No. 101211118.


\bibliography{aaai2026}

\end{document}